\documentclass[10pt,twocolumn,letterpaper]{article}

\usepackage[pagenumbers]{cvpr} 

\usepackage{url}
\usepackage{graphicx}
\usepackage{booktabs}
\usepackage{color, colortbl}
\usepackage{amssymb}
\usepackage{adjustbox}
\usepackage{algorithm}
\usepackage{algpseudocode}
\usepackage{amsthm}
\usepackage{bm}
\usepackage{tikz}
\usepackage{xcolor}
\usepackage{multirow}
\definecolor{LightCyan}{rgb}{0.88,1,1}
\usepackage{amsmath}
\usepackage[misc]{ifsym}

\makeatletter
\providecommand{\@LN}[2]{}
\makeatother
\definecolor{cvprblue}{rgb}{0.21,0.49,0.74}
\usepackage[pagebackref,breaklinks,colorlinks,allcolors=cvprblue]{hyperref}
\def\paperID{3900} 
\def\confName{CVPR}
\def\confYear{2025}

\title{$\bm{\mathcal{FOCUS}}$: Knowledge-enhanced Adaptive Visual Compression for Few-shot Whole Slide Image Classification
\vspace{-0.4cm}}
\author{Zhengrui Guo\textsuperscript{1,2}, Conghao Xiong\textsuperscript{3}, Jiabo Ma\textsuperscript{1}, Qichen Sun\textsuperscript{4}, Lishuang Feng\textsuperscript{2}, Jinzhuo Wang\textsuperscript{4},\\
and Hao Chen\textsuperscript{1(\Letter)}\\ 
\textsuperscript{1}HKUST, \textsuperscript{2}BICI, \textsuperscript{3}CUHK, \textsuperscript{4}Peking University\\
{\tt\small zguobc@connect.ust.hk,~jhc@cse.ust.hk}
\vspace{-0.4cm}}

\begin{document}
\maketitle

\begin{abstract}
Few-shot learning presents a critical solution for cancer diagnosis in computational pathology (CPath), addressing fundamental limitations in data availability, particularly the scarcity of expert annotations and patient privacy constraints. A key challenge in this paradigm stems from the inherent disparity between the limited training set of whole slide images (WSIs) and the enormous number of contained patches, where a significant portion of these patches lacks diagnostically relevant information, potentially diluting the model's ability to learn and focus on critical diagnostic features. While recent works attempt to address this by incorporating additional knowledge, several crucial gaps hinder further progress: (1) despite the emergence of powerful pathology foundation models (FMs), their potential remains largely untapped, with most approaches limiting their use to basic feature extraction; (2) current language guidance mechanisms attempt to align text prompts with vast numbers of WSI patches all at once, struggling to leverage rich pathological semantic information. To this end, we introduce the knowledge-enhanced adaptive visual compression framework, dubbed \textbf{FOCUS}, which uniquely combines pathology FMs with language prior knowledge to enable a focused analysis of diagnostically relevant regions by prioritizing discriminative WSI patches. Our approach implements a progressive three-stage compression strategy: we first leverage FMs for global visual redundancy elimination, and integrate compressed features with language prompts for semantic relevance assessment, then perform neighbor-aware visual token filtering while preserving spatial coherence. Extensive experiments on pathological datasets spanning breast, lung, and ovarian cancers demonstrate its superior performance in few-shot pathology diagnosis. Codes are available at \url{https://github.com/dddavid4real/FOCUS}.
\end{abstract}
\vspace{-0.2cm}
\section{Introduction}\label{sec:intro}
Traditional pathology has long relied on microscopic examination of tissue specimens for disease diagnosis, particularly in cancer detection \cite{van2021deep}. The digitization of glass slides into whole slide images (WSIs) has transformed this field, enabling computational analysis of these high-resolution images \cite{niazi2019digital}. This has given rise to computational pathology (CPath), where computational methods are introduced to automatically analyze WSIs for disease diagnosis \cite{cui2021artificial}. However, the ultra-high resolution nature of WSIs (often exceeding $100,000 \times 100,000$ pixels) and the fact that diagnostic patterns may be scattered across different tissue regions present unique challenges, naturally leading to the development of multi-instance learning (MIL), a weakly-supervised learning paradigm \cite{campanella2019clinical,lu2021data,shao2021transmil,zhang2022dtfd,xu2023multimodal,zhou2023cross,zhang2024prototypical,yang2024mambamil}.

MIL for slide-level diagnosis and prognosis typically follows three steps: first, the tissue regions (foreground) of WSI are segmented into smaller patches; second, these patches are encoded into feature representations using pre-trained feature extractors; finally, an aggregation mechanism combines these features to generate slide-level representations for analysis \cite{jin2023label}. While MIL framework has been proven highly effective in the field of CPath \cite{ilse2018attention,campanella2019clinical,chikontwe2020multiple,lu2021data,shao2021transmil,li2021dt,xiang2022dsnet,hou2022h,zheng2022graph,zhang2022dtfd,wang2022scl,yu2023prototypical,lu2023visual,lin2023interventional,li2024dynamic,yang2024mambamil,guo2024histgen,chen2020pathomic,song2024multimodal,chen2022pan,jaume2024modeling}, training current approaches typically demands large annotated datasets that are labor-intensive and time-consuming to obtain, particularly for rare pathological conditions \cite{hosseini2024computational}. These limitations underscore the pressing need for a learning mechanism that can effectively learn from limited examples while maintaining the advantages of traditional MIL approaches, potentially revolutionizing how CPath models are deployed in resource-constrained settings \cite{yang2022towards,song2023comprehensive}. Meanwhile, although MIL methods are designed to handle the inherent sparsity of diagnostic information in WSIs, where only a small fraction of patches contains relevant features, the few-shot setting introduces additional challenges. With limited training examples, models must learn to identify and leverage discriminative patterns from a drastically reduced set of WSIs, each containing thousands of patches with varying degrees of diagnostic relevance. To this end, the concept of few-shot weakly-supervised learning (FSWL) is first proposed by Qu \textit{et al.} \cite{qu2024rise}, aiming at addressing these challenges via a data-efficient manner. Building upon this initial framework, several follow-up studies have emerged in the field of CPath, proposing specialized MIL frameworks for FSWL scenarios \cite{shi2024vila, qu2024pathology, han2024mscpt, fu2024fast}. To achieve precise pathological diagnosis with limited training data, most existing approaches attempt to incorporate additional prior knowledge, including language-guided pathology description \cite{qu2024rise, shi2024vila,han2024mscpt} and additional visual slides as exemplars \cite{qu2024pathology}. While these explorations have demonstrated promising diagnosis results, they still face several key limitations:

\textbf{First, the potential of pathology foundation models (FMs) remains largely unexplored.} Modern pathology FMs have demonstrated remarkable abilities in understanding tissue architecture, cellular relationships, and morphological patterns across diverse pathological conditions \cite{ma2024towards,chen2024towards,xu2024whole,vorontsov2024foundation,wang2024pathology,lu2024visual,huang2023visual,xu2024multimodal}. Despite the significant advancements brought by pathology FMs to the field of CPath, most existing MIL frameworks either neglect to utilize them or simply employ them for WSI patch feature extraction. This limited utilization overlooks the rich semantic understanding and contextual learning capabilities that FMs possess through their pre-training on vast pathology datasets comprising billions of images.

\textbf{Second, language prompt integration mechanisms in existing FSWL frameworks fail to adequately capture complex pathological semantics.} Current FSWL approaches incorporate language prompts containing pathological prior knowledge to guide patch feature aggregation \cite{qu2024rise,shi2024vila, qu2024pathology, han2024mscpt, fu2024fast}. However, these approaches attempt to align text features with all patch features simultaneously, which proves problematic given that only a small subset of WSI patches typically contain diagnostically significant patterns relevant to the text prompt.
Such indiscriminate processing could potentially dilute the model's focus to diagnostically relevant regions. Therefore, identifying and prioritizing informative patches prior to language-guided aggregation emerges as a crucial consideration.

To address the above limitations and advance few-shot whole slide image analysis, we introduce the knowledge-enhanced adaptive visual compression framework, named \textbf{FOCUS}. The core idea of FOCUS lies in its progressive and adaptive compression of visual tokens in the patch sequence by leveraging pathology FMs and language prior knowledge, enabling the model to focus on diagnostically relevant regions. 
Our framework implements a three-stage visual compression strategy: (1) we introduce a global visual redundancy removal module that leverages discriminative features from FMs through sliding-window similarity measurements, enabling efficient coarse-grained filtering of redundant tokens; (2) a knowledge-enhanced adaptive visual token compression mechanism is designed to adaptively prioritize visual tokens based on their semantic relevance to textual pathology descriptions; and (3) we incorporate a sequential visual token compression module that processes tokens in pair-wise order and preserves dissimilar tokens based on cosine similarity thresholding, maintaining spatial continuity while eliminating local redundancies. This adaptive strategy directly addresses a key challenge in FSWL scenarios, \textit{i.e.}, the scarcity of training samples makes models especially vulnerable to interference from irrelevant tokens, and an excess of irrelevant tokens can potentially dilute the model's attention to critical diagnostic features. Our main contributions are summarized as follows:

\begin{itemize}
    \item We explores the untapped potential of pathology foundation models for few-shot weakly-supervised learning tasks, a direction largely overlooked in existing works. We move beyond using FMs as mere feature extractors and showcase how their rich representational capabilities can be harnessed.
    \item We introduce FOCUS, a knowledge-enhanced adaptive visual compression framework for precise diagnosis of gigapixel WSIs with limited training data. FOCUS leverages knowledge from pathology FMs and language prior to progressively concentrate the model's attention on diagnostically relevant regions, pushing the boundaries of CPath in data-scarce settings.
    \item FOCUS achieves superior or comparable performance against state-of-the-art (SOTA) methods in extensive evaluations across diverse cancer datasets under few-shot settings, demonstrating its effectiveness in capturing diagnostically relevant features in data-efficient scenarios. Through comprehensive ablation studies, we systematically analyze the contributions of each component, including the three-stage compression strategy, the choice of pathology FMs, and the design of language prompts, validating the effectiveness of our proposed architecture.
\end{itemize}
\section{Related Works}
\label{sec:related_works}

\noindent\textbf{Multiple instance learning in CPath.} Due to the gigapixel size of WSIs and GPU memory constraints \cite{araujo2019computing}, the MIL framework has become the predominant approach for WSI analysis. In recent years, MIL-based methods have demonstrated remarkable success in weakly-supervised WSI diagnosis and prognosis \cite{ilse2018attention,campanella2019clinical,chikontwe2020multiple,lu2021data,shao2021transmil,li2021dt,xiang2022dsnet,hou2022h,zheng2022graph,zhang2022dtfd,wang2022scl,yu2023prototypical,lu2023visual,lin2023interventional,li2024dynamic,song2024morphological,yang2024mambamil,guo2024histgen,xiong2023diagnose,xiong2024mome,xiong2024takt,guo2025context,du2025beyond,zheng2025diffusion}. MIL methods typically comprise two key components: a feature extractor and an aggregator. The feature extractor (usually pre-trained) projects WSI patches into a latent space to obtain their representations, then the aggregator combines these patch-level representations to create a slide-level representation for diagnosis or prognosis. Traditional approaches rely on simple, non-parametric aggregation methods such as max and mean pooling operators \cite{campanella2019clinical}. Subsequent research has focused on developing more sophisticated aggregation algorithms to better identify and capture critical diagnostic patterns distributed across thousands of patches. Ilse \textit{et al.} \cite{ilse2018attention} introduce ABMIL, an attention-based model that leverages information from all patches by employing a side network to compute attention scores, which serve as weights during aggregation. DS-MIL \cite{li2021dual} exploits the hierarchical nature of WSIs through a dual-stream architecture that aggregates multi-scale features extracted from different magnification levels. TransMIL \cite{shao2021transmil} adapts the Transformer architecture to WSI analysis, enabling effective modeling of long-range correlations between patches. DTFD-MIL \cite{zhang2022dtfd} further extends the existing MIL paradigms by introducing the concept of pseudo-bag and developing a double-tier framework for more effective feature utilization. WiKG \cite{li2024dynamic} represents a WSI as a knowledge graph, treating cropped patches as nodes and leveraging head-to-tail patch embeddings to generate dynamic graph representations for WSI analysis. Further, several studies have incorporated pathology image captions \cite{lu2023visual} and diagnosis reports \cite{guo2024histgen} to enhance WSI interpretation and diagnostic accuracy. Although these methods have advanced the field significantly, they typically require large annotated datasets for training, making them less suitable for scenarios with limited data availability, which is a common challenge in clinical settings. Our FOCUS framework addresses this limitation by introducing knowledge-enhanced adaptive visual compression to enable effective learning from just a few labeled examples.

\noindent\textbf{Foundation models in CPath.} In this field of CPath, an ever-increasing number of foundation models have been proposed to improve diagnostic accuracy and prognostic assessment fundamentally by leveraging vast amounts of pathology data \cite{ma2024towards,chen2024towards,xu2024whole,vorontsov2024foundation,wang2024pathology,lu2024visual,huang2023visual,xu2024multimodal}. These efforts involve vision-only foundation models such as GPFM \cite{ma2024towards}, UNI \cite{chen2024towards}, Prov-Gigapath \cite{xu2024whole}, Virchow \cite{vorontsov2024foundation}, and CHIEF \cite{wang2024pathology}, as well as multimodal models that integrate WSIs with pathology reports or genomic profiles, including CONCH \cite{lu2024visual}, PLIP \cite{huang2023visual}, and mSTAR \cite{xu2024multimodal}. While foundation models commonly serve as feature extractors in MIL frameworks to provide robust feature initialization, we argue that the rich representational capabilities of these models remain underexplored. Specifically, we leverage the patch-level features extracted by foundation models to compute inter-patch similarities, enabling adaptive patch reduction that focuses the model's attention on the most diagnostically relevant regions. This similarity-thresholding-based sequence compression not only enhances computational efficiency but also improves few-shot learning performance by prioritizing informative patches. 

\noindent\textbf{Few-shot weakly-supervised learning in CPath.} TOP \cite{qu2024rise} firstly introduces the concept of FSWL, which considers a limited number of WSIs (usually $1$, $4$, $8$, or $16$ per class) as training samples, aiming at improving model capabilities in data-scarce scenarios. Specifically, this work develops a two-level prompt learning MIL framework guided by pathology language prior knowledge. Following this paradigm, more sophisticated frameworks have been proposed to capture crucial diagnostic information with limited training samples \cite{shi2024vila, qu2024pathology, han2024mscpt, fu2024fast}. For instance, ViLa-MIL \cite{shi2024vila} proposes a dual-scale prompt learning framework to aggregate high- and low-resolution slide-level features with descriptive texts. PEMP \cite{qu2024pathology} enhances pathology feature understanding in data-limited scenarios by incorporating task-specific image examples. MSCPT \cite{han2024mscpt} employs a graph prompt tuning module with graph convolutional networks to capture essential contextual information within WSIs. However, their explorations remain preliminary. 
Moreover, existing works impose impractical requirements such as paired multi-resolution images \cite{shi2024vila,han2024mscpt} or additional reference samples \cite{qu2024pathology}, limiting their clinical applicability. FOCUS overcomes these limitations by introducing a three-stage compression strategy that operates on standard single-resolution WSIs while leveraging both foundation models and language prompts for focused analysis of diagnostically relevant regions.
\section{Methodology}\label{sec:method}
\begin{figure*}[t]
  \centering
  \includegraphics[width=\textwidth]{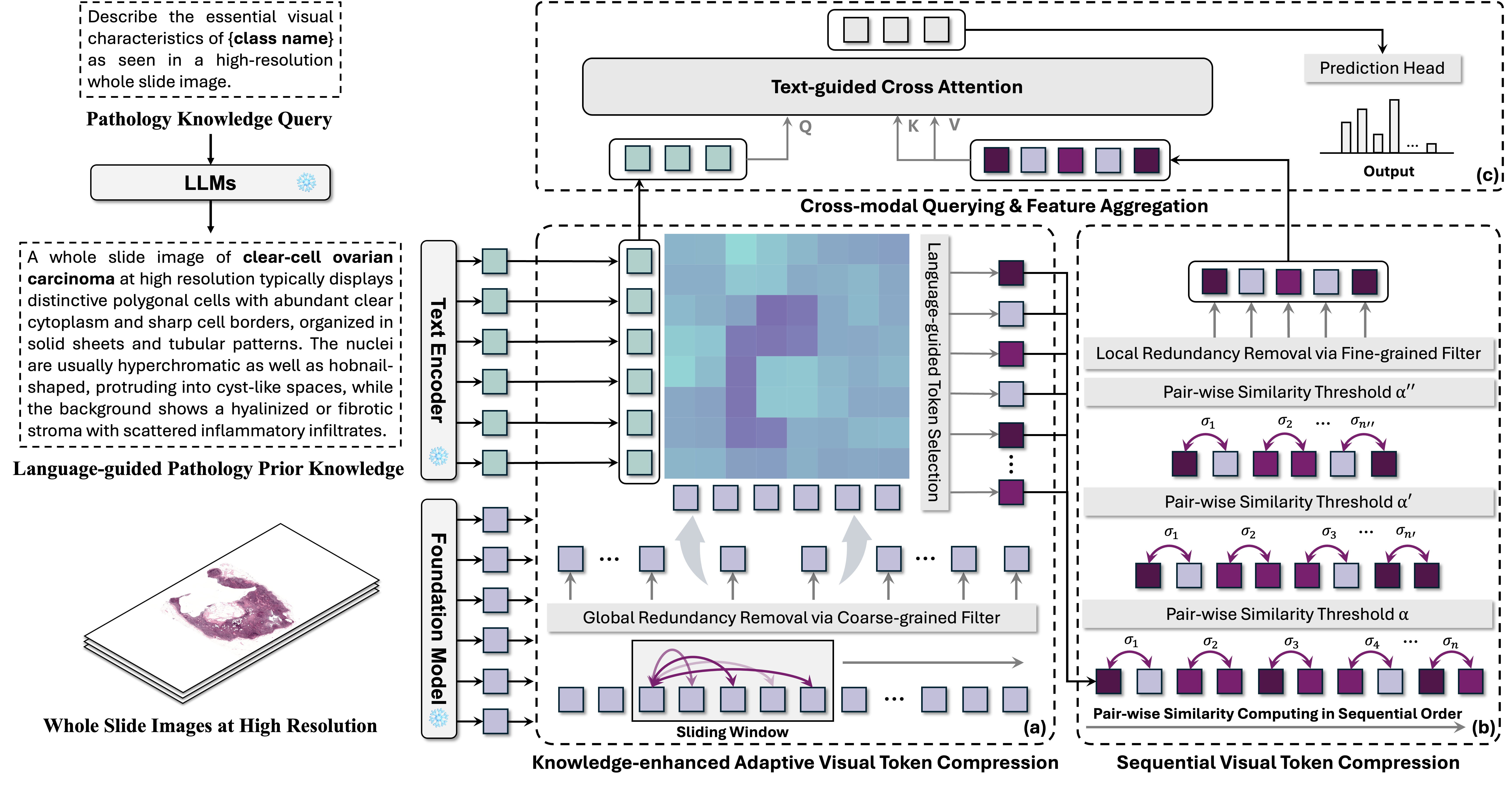}
  \caption{Illustration of the proposed \textbf{FOCUS} framework, which consists of three key components: (a) A knowledge-enhanced adaptive visual token compression module that performs global redundancy removal via FM features and language prior-guided visual token prioritization, (b) A sequential visual token compression module that performs pair-wise similarity thresholding to eliminate local redundancies while preserving spatial coherence, where $\sigma_i$ denotes the pairwise similarity (cosine similarity) between adjacent patches, and (c) A cross-modal aggregation module that combines the compressed visual features with pathology knowledge prompts for final prediction.} 
  \vspace{-0.2cm}
  \label{fig:methodology}
\end{figure*}

\subsection{Overview and Problem Formulation}\label{sec:3.1}
This work introduces the knowledge-enhanced adaptive visual compression framework for few-shot whole slide image analysis, dubbed FOCUS, as illustrated in Figure \ref{fig:methodology}. FOCUS comprises three key components: the knowledge-enhanced adaptive visual token compression module, the sequential visual token compression module, and the cross-modal aggregation module. These components work in concert to achieve adaptive visual token compression by leveraging pathology FMs and domain-specific prior knowledge. The overall workflow of FOCUS is further elaborated by pseudo-code (see Algorithm \ref{alg:lavic-mil} in Supplementary Material).

\subsubsection{WSI Bag Construction}
Formally, let $X \in \mathbb{R}^{H \times W \times 3}$ be a whole slide image (\textit{i.e.}, a bag). Due to the gigapixel resolution of WSIs, direct processing is computationally infeasible. Following standard practice, we first divide the foreground of $X$ into $N$ non-overlapping patches $X=\{x_1, x_2, ..., x_{N}\}$ where each $x_i \in \mathbb{R}^{h \times w \times 3}$. Each WSI $X$ is assigned with a slide-level label $Y\in \{0, 1, \ldots, S\}$, where $S$ is the number of classes. The label for each patch (instance) inside the bag is unavailable. These patches are projected through a pre-trained model $f(\cdot)$ (pathology foundation models in our implementation) to obtain a bag of patch-level representations:
\begin{equation}\label{eq:feautre_extraction}
    \mathbf{B}=\{f(x_i): x_i \in X\}=\{\mathbf{b}_i\}_{i=1}^{N},
\end{equation}
where $\mathbf{b}_i \in \mathbb{R}^d$ and $d$ is the feature dimension. Unlike multi-scale approaches, our framework operates on a single WSI scale (high-resolution scale in our implementation), simplifying the input requirements. Note that in the FSWL setting, the model receives only $K$ labeled examples per class (also known as \textit{$K$-shots}) for training, while evaluated on the entire testing set. The typical choices of $K$ are $1,2,4,8,16, 32$. 

\subsubsection{Pathology Knowledge Prompt Generation}
To bridge the semantic gap between complex visual patterns and their diagnostic significance in data-limited scenarios, we incorporate language prompts containing standardized pathology terminology and expert knowledge (as shown in Figure \ref{fig:methodology}). For each cancer subtype, we formulate a pathology knowledge query: \textit{Please describe the essential visual characteristics of $\{$class name$\}$ as seen in a high-resolution whole slide image}. We then leverage large language models (\textit{e.g.}, Claude, ChatGPT) to generate a comprehensive description that contains diagnostic criteria and domain-specific morphological feature characterization. For example, in characterizing the \textit{clear-cell ovarian carcinoma} subtype of ovarian cancer, the pathology description encapsulates expert knowledge with several key aspects, ranging from nuclear and cellular features (\textit{``hobnail-shaped nuclei"}) to architectural patterns (\textit{``solid sheets and tubular arrangements"}) and stromal characteristics (\textit{``hyalinized stroma with inflammatory infiltrates"}). This detailed description provides crucial prior knowledge to compensate for limited training data, effectively bridging the gap between visual features and diagnostic criteria. This description then serves as a prompt and is mapped through a text encoder $E_T(\cdot)$ to the embedding space to generate the text prompt embedding $\mathbf{T^P} \in \mathbb{R}^{t_1\times d}$.
Following the practice of existing works \cite{zhou2022learning,shi2024vila}, we augment the language prompts $\mathbf{T^P}$ by prepending learnable prompts $\mathbf{T^L}\in \mathbb{R}^{t_2\times d}$, which are optimized during training to capture implicit patterns and facilitate more flexible cross-modal feature alignment.

\subsection{Knowledge-enhanced Adaptive Compression}\label{sec:3.2}
As shown in Figure \ref{fig:methodology}(a), the knowledge-enhanced adaptive visual token compression module leverages both foundation models and pathology domain knowledge to conduct visual token compression and identify diagnostically relevant patches. This module implements a two-stage selection strategy: global redundancy removal through coarse-grained filtering via pathology FM features, followed by language-guided visual token prioritization.

\subsubsection{Global Redundancy Removal via Pathology FMs}\label{sec:3.2.1}
Given a bag of patch representations $\mathbf{B} = \{\mathbf{b}_i\}^{N}_{i=1}$ extracted by the pathology foundation model (CONCH \cite{lu2024visual} in our implementation), we first address the challenge of redundant visual information through a sliding window mechanism. For each $1$-dimensional window of size $w$, we compute normalized features and their pairwise similarities:
\begin{equation}\label{eq:fm}
    \hat{\mathbf{b}}_i = \frac{\mathbf{b}_i}{\|\mathbf{b}_i\|_2}, \quad S_{ij} = \hat{\mathbf{b}}_i \cdot \hat{\mathbf{b}}_j, \quad \tau_g = \mu(S) + \sigma(S)
\end{equation}
where $\tau_g$ represents a dynamic threshold computed from the mean $\mu(S)$ and standard deviation $\sigma(S)$ of similarities within the window. Patches are considered redundant if their mean similarity exceeds $\tau_g$: $R_i = \frac{1}{w}\sum_{j=1}^w S_{ij} > \tau_g$. Operating as a coarse-grained filter, this module provides a broad-stroke assessment of redundancy across the WSI, serving as an initial screening mechanism before more sophisticated compression processes.
Sliding window ensures that redundancy is assessed within broader tissue contexts rather than just between neighboring patches, while the dynamic thresholding adapts to feature distributions within each window, making it particularly effective for eliminating large-scale repetitive patterns that contribute limited diagnostic information.

\subsubsection{Language-guided Visual Token Prioritization}\label{sec:3.2.2}
Following global redundancy removal, we introduce a language-guided visual token prioritization mechanism to align the remaining visual tokens with pathology semantic information. Given the concatenated text embeddings $\mathbf{T} = [\mathbf{T^L}; \mathbf{T^P}] \in \mathbb{R}^{(t_1+t_2) \times d}$, where $\mathbf{T^L}$ represents learnable prompts and $\mathbf{T^P}$ denotes pathology knowledge prompts, we compute cross-modal attention scores:
\begin{equation}\label{eq:att_score}
    \mathbf{A} = \text{softmax}(\frac{(\mathbf{T}W_q)(\mathbf{B}W_k)^\top}{\sqrt{d}}), ~~r_i = \frac{1}{t_1+t_2}\sum_{j=1}^{t_1+t_2} \mathbf{A}_{ji},
\end{equation}
where $W_q, W_k \in \mathbb{R}^{d \times d}$ are learnable projection matrices and $r_i$ represents each token's relevance score. The number of tokens to select (denoted by $k$) and final selected sequence (denoted by $\mathbf{B}_s$) are determined by:
\begin{equation}\label{eq:text_select}
    k = \min(M_{max}, \gamma N'), \quad \mathbf{B}_s = \{\mathbf{b}_i | \text{rank}(r_i) \leq k\},
\end{equation}
where $M_{max}$ is a predefined maximum sequence length to ensure computational efficiency, $N'$ represents the current sequence length, $\gamma \in (0,1)$ is the predefined compression ratio, and $\mathbf{B}_s$ denotes the selected top-$k$ tokens based on their relevance rankings. By leveraging domain knowledge encoded in language prompts, the module effectively prioritizes patches exhibiting characteristics typically associated with specific cancer subtypes. Through cross-modal relevance computing, the framework captures subtle correlations between visual patterns and pathological descriptions, producing selected tokens that are both visually distinctive and clinically meaningful.

\subsection{Sequential Visual Token Compression}\label{sec:3.3}
As illustrated in Figure \ref{fig:methodology}(b), we introduce a sequential visual token compression module that processes tokens in sequential order to eliminate local redundancies while maintaining relative spatial structure. This module implements a multi-stage compression strategy with progressively increasing similarity thresholds. Given a sequence of selected tokens $\mathbf{B}_s = \{\mathbf{b}_1, \mathbf{b}_2, \ldots, \mathbf{b}_k\}$, at each stage $i$, we compute the cosine similarity between consecutive tokens:
\begin{equation}\label{eq:sequential_compression}
    s_{j,j+1} = \frac{\mathbf{b}^{l\top}_j \mathbf{b}_{j+1}}{\|\mathbf{b}_j\|_2 \cdot \|\mathbf{b}_{j+1}\|_2}, \quad j \in \{1,\ldots,k-1\}
\end{equation}
The compression follows a threshold-based rule with increasing thresholds $\theta_i = \theta_{\text{base}} + i\Delta\theta$ to generate binary masks:
\begin{equation}\label{eq:threshold}
    \text{mask}_{j} = \begin{cases}
        1, & \text{if } \min(s_{j-1,j}, s_{j,j+1}) < \theta_i \\
        0, & \text{otherwise}
    \end{cases}
\end{equation}
where $\Delta\theta$ controls the increase in filtering intensity across stages. The sequence is progressively compressed as $\mathbf{B}_{i+1} = \{\mathbf{b}_j \in \mathbf{B}_i : \text{mask}_j = 1\}$, with the final compressed sequence $\mathbf{B}_c$ containing tokens that maintain distinctiveness through all compression stages.

\subsection{Cross-modal Aggregation for Prediction}\label{sec:3.4}
Following the compression stages, we employ a cross-modal aggregation module, illustrated in Figure \ref{fig:methodology}(c), to combine the compressed visual tokens with pathological semantic information for final prediction. Given the compressed sequence $\mathbf{B}_c \in \mathbb{R}^{N_c \times d}$ and text embeddings $\mathbf{T} = [\mathbf{T^L}; \mathbf{T^P}]$, we first compute text-guided aggregation through a multi-head attention mechanism, then the outputs from all heads are concatenated and normalized:
\begin{equation}\label{eq:cross_attn}
    \text{Head}_i = \text{softmax}(\frac{\mathbf{Q}W^i_q(\mathbf{K}W^i_k)^\top}{\sqrt{d_k}})\mathbf{V}W^i_v,
\end{equation}
\begin{equation}\label{eq:concat_head}
    \mathbf{O} = \text{LayerNorm}(\text{Concat}(\text{Head}_1, ..., \text{Head}_h)W_o),
\end{equation}
where $\mathbf{Q} = \mathbf{T}W_q$, $\mathbf{K} = \mathbf{B}_cW_k$, $\mathbf{V} = \mathbf{B}_cW_v$ are query, key, and value projections, respectively, and $d_k$ is the dimension per head. The final prediction is obtained through a classification head:
\begin{equation}\label{eq:prediction}
    P(Y|\mathbf{B}_c, \mathbf{T}) = \text{softmax}(W_c \mathbf{O} + \beta_c),
\end{equation}
where $W_c$ and $\beta_c$ are learnable parameters. 

\begingroup
\setlength{\tabcolsep}{4pt} 
\renewcommand{\arraystretch}{1.5} 
\begin{table*}[ht]
\centering
\caption{Few-shot weakly-supervised learning results on TCGA-NSCLC, CAMELYON, and UBC-OCEAN datasets under 4-shot, 8-shot, and 16-shot settings. The best results are in bold, and the second-best results are underlined.}
\begin{adjustbox}{width=\textwidth}
\scalebox{0.8}{
\begin{tabular}{c|cccccccccccc}
\toprule
\toprule
\multirow{2}{*}{\textbf{Dataset}} & \multicolumn{1}{c}{\multirow{2}{*}{\textbf{Methods}}} & \multicolumn{3}{c}{\textbf{4-shot}} &  &\multicolumn{3}{c}{\textbf{8-shot}} &  & \multicolumn{3}{c}{\textbf{16-shot}} \\ 
\cline{3-5} \cline{7-9} \cline{11-13}
& & Balanced ACC & AUC & F1 Score & & Balanced ACC & AUC & F1 Score & & Balanced ACC & AUC & F1 Score\\ 
\midrule
\midrule
\multirow{10}{*}{\rotatebox[origin=c]{90}{\textbf{TCGA-NSCLC (2 classes)}}} 
& Mean Pooling &$0.722_{\pm 0.059}$& $0.801_{\pm 0.068}$& $0.719_{\pm 0.060}$ & & $0.779_{\pm 0.054}$&$0.867_{\pm 0.051}$ &$0.777_{\pm 0.055}$ & &$0.819_{\pm 0.042}$ & $0.899_{\pm 0.040}$& $0.818_{\pm 0.042}$\\
& Max Pooling  & $0.755_{\pm 0.053}$& $0.849_{\pm 0.061}$&$0.753_{\pm 0.053}$ & & $0.829_{\pm 0.062}$ & \underline{$0.917_{\pm 0.050}$} & $0.828_{\pm 0.062}$&  & $0.863_{\pm 0.048}$ &$0.940_{\pm 0.042}$ & $0.862_{\pm 0.048}$\\
& Attentive Pooling   &  $0.782_{\pm 0.046}$  &  $0.856_{\pm 0.040}$  &  $0.781_{\pm 0.047}$ &  &$0.825_{\pm 0.029}$ & $0.904_{\pm 0.028}$& $0.823_{\pm 0.029}$ & & $0.852_{\pm 0.036}$ &$0.930_{\pm 0.034}$ & $0.852_{\pm 0.036}$\\
& TransMIL \cite{shao2021transmil}   & $0.745_{\pm 0.058}$ & $0.811_{\pm 0.067}$ & $0.743_{\pm 0.048}$ & & $0.807_{\pm 0.061}$ & $0.889_{\pm 0.058}$ & $0.806_{\pm 0.061}$ & &$0.872_{\pm 0.047}$ & $0.937_{\pm 0.036}$& $0.872_{\pm 0.047}$ \\
& DS-MIL \cite{li2021dual}   & $0.769_{\pm 0.050}$ & $0.837_{\pm 0.065}$ & $0.766_{\pm 0.052}$& & \underline{$0.844_{\pm 0.050}$} & $0.911_{\pm 0.045}$ & \underline{$0.843_{\pm 0.051}$} & &$0.879_{\pm 0.034}$ &$0.955_{\pm 0.021}$ &$0.879_{\pm 0.035}$\\
& DTFD-MIL \cite{zhang2022dtfd}   & $0.760_{\pm 0.066}$ & $0.842_{\pm 0.062}$ & $0.759_{\pm 0.067}$& & $0.839_{\pm 0.064}$ & $0.916_{\pm 0.051}$ & $0.838_{\pm 0.065}$ & & \underline{$0.887_{\pm 0.032}$} & \underline{$0.958_{\pm 0.022}$} & \underline{$0.886_{\pm 0.032}$}\\
& WiKG-MIL \cite{li2024dynamic} & $0.745_{\pm 0.064}$ & $0.816_{\pm 0.077}$ & $0.744_{\pm 0.064}$&  & $0.814_{\pm 0.046}$ & $0.897_{\pm 0.043}$ & $0.813_{\pm 0.046}$&  & $0.845_{\pm 0.040}$ & $0.917_{\pm 0.037}$ & $0.844_{\pm 0.040}$ \\
& TOP-MIL \cite{qu2024rise}   & $0.574_{\pm 0.087}$ & $0.611_{\pm 0.114}$ & $0.545_{\pm 0.115}$ & & $0.640_{\pm 0.109}$ & $0.678_{\pm 0.157}$ & $0.611_{\pm 0.136}$&  & $0.682_{\pm 0.107}$ & $0.742_{\pm 0.141}$& $0.671_{\pm 0.116}$\\
& ViLa-MIL \cite{shi2024vila} &\underline{$0.807_{\pm 0.043}$} & \underline{$0.885_{\pm 0.041}$} & \underline{$0.807_{\pm 0.044}$} & & $0.832_{\pm 0.060}$ & $0.905_{\pm 0.054}$ & $0.830_{\pm 0.062}$ & &$0.862_{\pm 0.033}$ & $0.942_{\pm 0.026}$ & $0.861_{\pm 0.031}$\\
& \cellcolor{gray!20!white} FOCUS (Ours)  & \cellcolor{gray!20!white} $\bm{0.819_{\pm 0.044}}$ & \cellcolor{gray!20!white} $\bm{0.915_{\pm 0.057}}$ & \cellcolor{gray!20!white} $\bm{0.819_{\pm 0.067}}$ &\cellcolor{gray!20!white} & \cellcolor{gray!20!white} $\bm{0.855_{\pm 0.058}}$& \cellcolor{gray!20!white} $\bm{0.940_{\pm 0.037}}$ & \cellcolor{gray!20!white} $\bm{0.853_{\pm 0.077}}$ &\cellcolor{gray!20!white} &\cellcolor{gray!20!white} $\bm{0.898_{\pm 0.026}}$ &\cellcolor{gray!20!white} $\bm{0.972_{\pm 0.017}}$&\cellcolor{gray!20!white} $\bm{0.897_{\pm 0.024}}$ \\
\midrule
\multirow{10}{*}{\rotatebox[origin=c]{90}{\textbf{CAMELYON (2 classes)}}} 
& Mean Pooling  &  $0.638_{\pm 0.035}$  & $0.626_{\pm 0.060}$ & $0.494_{\pm 0.063}$&  & $0.604_{\pm 0.057}$ & $0.606_{\pm 0.072}$ & $0.508_{\pm 0.061}$ & & $0.658_{\pm 0.045}$ & $0.665_{\pm 0.047}$&$0.603_{\pm 0.043}$\\
& Max Pooling     & $0.629_{\pm 0.101}$&  $0.632_{\pm 0.119}$ & $0.553_{\pm 0.132}$ & & $0.684_{\pm 0.125}$ &  $0.734_{\pm 0.151}$  &  $0.671_{\pm 0.131}$& & $0.817_{\pm 0.124}$ &$0.864_{\pm 0.146}$ &$0.807_{\pm 0.128}$ \\
& Attentive Pooling   &$0.611_{\pm 0.052}$& $0.614_{\pm 0.080}$ &  $0.546_{\pm 0.075}$ &  & $0.656_{\pm 0.054}$ & $0.682_{\pm 0.062}$ & $0.614_{\pm 0.042}$ & & $0.720_{\pm 0.049}$ &$0.738_{\pm 0.040}$ &$0.690_{\pm 0.045}$\\
& TransMIL \cite{shao2021transmil}   & $0.626_{\pm 0.036}$ & $0.587_{\pm 0.095}$ & $0.505_{\pm 0.074}$& & $0.658_{\pm 0.051}$  & $0.655_{\pm 0.073}$  & $0.610_{\pm 0.066}$ & & $0.786_{\pm 0.066}$ & $0.827_{\pm 0.088}$ & $0.768_{\pm 0.070}$ \\
& DS-MIL \cite{li2021dual}   & $0.643_{\pm 0.074}$& $0.639_{\pm 0.087}$ & \underline{$0.608_{\pm 0.076}$} && $0.782_{\pm 0.064}$ & $0.794_{\pm 0.066}$ & $0.756_{\pm 0.072}$ & & $0.849_{\pm 0.037}$ &$0.874_{\pm 0.046}$ &$0.830_{\pm 0.044}$ \\
& DTFD-MIL \cite{zhang2022dtfd}   & $0.609_{\pm 0.103}$ & $0.619_{\pm 0.117}$ & $0.587_{\pm 0.106}$ &&$0.769_{\pm 0.094}$ & \underline{$0.845_{\pm 0.094}$} & $0.759_{\pm 0.091}$ &&  $0.889_{\pm 0.029}$ & \underline{$0.934_{\pm 0.030}$} & $0.883_{\pm 0.030}$ \\
& WiKG-MIL \cite{li2024dynamic}   &  $0.637_{\pm 0.050}$ & $0.590_{\pm 0.109}$ & $0.479_{\pm 0.118}$ && $0.713_{\pm 0.071}$ & $0.721_{\pm 0.092}$ & $0.659_{\pm 0.105}$ && $0.890_{\pm 0.028}$ &$0.925_{\pm 0.032}$ & $0.883_{\pm 0.029}$\\
& TOP-MIL \cite{qu2024rise}   & $0.646_{\pm 0.082}$ & $0.605_{\pm 0.151}$ & $0.527_{\pm 0.133}$ & &  $0.669_{\pm 0.140}$ & $0.663_{\pm 0.174}$ &$0.635_{\pm 0.155}$& & $0.739_{\pm 0.119}$ &$0.751_{\pm 0.166}$ & $0.697_{\pm 0.161}$ \\
& ViLa-MIL \cite{shi2024vila} & \underline{$0.658_{\pm 0.036}$} & \underline{$0.666_{\pm 0.063}$} & $0.538_{\pm 0.123}$ && \underline{$0.800_{\pm 0.051}$} & $0.805_{\pm 0.084}$ &\underline{$0.765_{\pm 0.071}$} && $\bm{0.904_{\pm 0.030}}$ & $0.923_{\pm 0.031}$ & $\bm{0.895_{\pm 0.034}}$\\
& \cellcolor{gray!20!white} FOCUS (Ours)  & \cellcolor{gray!20!white} $\bm{0.701_{\pm 0.082}}$ & \cellcolor{gray!20!white} $\bm{0.700_{\pm 0.021}}$& \cellcolor{gray!20!white} $\bm{0.631_{\pm 0.046}}$ &\cellcolor{gray!20!white} & \cellcolor{gray!20!white}$\bm{0.833_{\pm 0.074}}$ & \cellcolor{gray!20!white} $\bm{0.856_{\pm 0.095}}$ & \cellcolor{gray!20!white} $\bm{0.807_{\pm 0.084}}$ &\cellcolor{gray!20!white} &\cellcolor{gray!20!white}\underline{ $0.900_{\pm 0.033}$} &\cellcolor{gray!20!white} $\bm{0.943_{\pm 0.027}}$&\cellcolor{gray!20!white} \underline{$0.890_{\pm 0.035}$}\\
\midrule
\multirow{10}{*}{\rotatebox[origin=c]{90}{\textbf{UBC-OCEAN (5 classes)}}} 
& Mean Pooling  &$0.656_{\pm 0.059}$  & $0.888_{\pm 0.031}$ &$0.603_{\pm 0.048}$ &&$0.702_{\pm 0.052}$ &$0.926_{\pm 0.028}$ &$0.673_{\pm 0.049}$ && $0.783_{\pm 0.033}$& $0.945_{\pm 0.020}$& $0.702_{\pm 0.058}$\\
& Max Pooling  &$0.640_{\pm 0.072}$  &$0.877_{\pm 0.040}$  &$0.601_{\pm 0.083}$ && $0.673_{\pm 0.103}$ & $0.899_{\pm 0.067}$ & $0.644_{\pm 0.094}$ && $0.820_{\pm 0.043}$&$0.949_{\pm 0.017}$ &$0.743_{\pm 0.088}$ \\
& Attentive Pooling  & $\underline{0.671_{\pm 0.080}}$ &  $0.898_{\pm 0.035}$  & $\underline{0.630_{\pm 0.077}}$ &&  $0.758_{\pm 0.055}$ & $0.935_{\pm 0.025}$  & $\underline{0.716_{\pm 0.057}}$ &&$0.804_{\pm 0.050}$ &$0.952_{\pm 0.021}$ & $0.715_{\pm 0.076}$\\
& TransMIL \cite{shao2021transmil}   & $0.619_{\pm 0.067}$ & $0.862_{\pm 0.046}$ & $0.550_{\pm 0.070}$ && $0.704_{\pm 0.068}$ & $0.917_{\pm 0.034}$ & $0.668_{\pm 0.079}$  && $0.795_{\pm 0.035}$ & $0.949_{\pm 0.011}$& $0.702_{\pm 0.062}$ \\
&  DS-MIL \cite{li2021dual}   & $0.646_{\pm 0.075}$ & $0.865_{\pm 0.058}$ & $0.589_{\pm 0.097}$ && $0.733_{\pm 0.075}$ & $0.925_{\pm 0.034}$ & $0.699_{\pm 0.077}$ && $0.825_{\pm 0.032}$& $\underline{0.956_{\pm 0.012}}$& $0.739_{\pm 0.048}$ \\
& DTFD-MIL \cite{zhang2022dtfd}  & $0.662_{\pm 0.089}$ & $\underline{0.901_{\pm 0.032}}$ & $0.626_{\pm 0.057}$ && $0.752_{\pm 0.055}$ & $0.928_{\pm 0.021}$ & $0.703_{\pm 0.073}$ && $\underline{0.854_{\pm 0.040}}$& $0.953_{\pm 0.023}$& $\underline{0.774_{\pm 0.068}}$  \\
& WiKG-MIL \cite{li2024dynamic}   & $0.633_{\pm 0.067}$ & $0.877_{\pm 0.031}$ & $0.585_{\pm 0.050}$ &&  $0.731_{\pm 0.060}$ & $0.918_{\pm 0.033}$ & $0.688_{\pm 0.072}$ && $0.831_{\pm 0.031}$ &$0.952_{\pm 0.015}$ & $0.757_{\pm 0.051}$ \\
& TOP-MIL \cite{qu2024rise}  & $0.619_{\pm 0.127}$ & $0.869_{\pm 0.044}$ & $0.584_{\pm 0.106}$ && $\underline{0.754_{\pm 0.066}}$ & $\underline{0.936_{\pm 0.020}}$ & $0.711_{\pm 0.070}$ && $0.846_{\pm 0.033}$& $\underline{0.956_{\pm 0.020}}$& $0.757_{\pm 0.060}$ \\
& ViLa-MIL \cite{shi2024vila} &$0.640_{\pm 0.051}$ & $0.869_{\pm 0.039}$ & $0.566_{\pm 0.068}$ && $0.750_{\pm 0.068}$ & $0.916_{\pm 0.029}$ & $0.704_{\pm 0.091}$ && $0.810_{\pm 0.055}$ & $0.939_{\pm 0.023}$ & $0.724_{\pm 0.065}$  \\
& \cellcolor{gray!20!white} FOCUS (Ours) & \cellcolor{gray!20!white} $\bm{0.704_{\pm 0.088}}$ & \cellcolor{gray!20!white} $\bm{0.911_{\pm 0.036}}$& \cellcolor{gray!20!white} $\bm{0.655_{\pm 0.094}}$& \cellcolor{gray!20!white} & \cellcolor{gray!20!white} $\bm{0.773_{\pm 0.054}}$& \cellcolor{gray!20!white} $\bm{0.945_{\pm 0.029}}$ & \cellcolor{gray!20!white} $\bm{0.735_{\pm 0.058}}$ & \cellcolor{gray!20!white}&\cellcolor{gray!20!white} $\bm{0.864_{\pm 0.032}}$ &\cellcolor{gray!20!white} $\bm{0.967_{\pm 0.015}}$ &\cellcolor{gray!20!white} $\bm{0.788_{\pm 0.051}}$ \\
\bottomrule
\bottomrule
\end{tabular}}
\end{adjustbox}
\label{tab:results_16}
\end{table*}
\endgroup

\subsection{Training Strategy}\label{sec:3.5}
The model is trained end-to-end using cross-entropy loss to minimize the negative log-likelihood of the ground truth labels:
\begin{equation}
    \mathcal{L} = -\log P(Y|\mathbf{B}_c, \mathbf{T}).
\end{equation}

\section{Experiments and Results}\label{sec:experiment}
\subsection{Experimental Settings}\label{sec:4.1}
\noindent \textbf{Datasets.} To demonstrate the effectiveness of our proposed framework FOCUS, we conduct extensive experiments on three publicly available pathology datasets, \textit{i.e.}, TCGA-NSCLC\footnote{https://portal.gdc.cancer.gov} \cite{tomczak2015review}, CAMELYON\footnote{https://camelyon16.grand-challenge.org/Data/}\textsuperscript{,}\footnote{https://camelyon17.grand-challenge.org/Data/}  \cite{bejnordi2017diagnostic,bandi2018detection}, and UBC-OCEAN\footnote{https://www.kaggle.com/competitions/UBC-OCEAN}, covering lung, breast, and ovarian cancers. TCGA-NSCLC comprises two lung cancer subtypes: lung adenocarcinoma (LUAD, 541 slides) and lung squamous cell carcinoma (LUSC, 512 slides). CAMELYON contains 577 normal slides and 341 slides with breast cancer metastasis. UBC-OCEAN encompasses five ovarian cancer subtypes: clear cell carcinoma (CC, 98 slides), endometrioid carcinoma (EC, 122 slides), high-grade serous carcinoma (HGSC, 221 slides), low-grade serous carcinoma (LGSC, 43 slides), and mucinous carcinoma (MC, 43 slides). Each dataset is split into train/validation/test sets with a ratio of 6:2:2, and we randomly sample $K$ samples per class under the $K$-shot FSWL setting ($K=4,8,16$ in our implementation).

\noindent \textbf{Implementation Details.}
We utlize CLAM toolset \cite{lu2021data} for WSI pre-processing, cropping the high-resolution ($20\times$ or $40\times$) WSIs into patches of $512\times 512$ pixels for feature extraction. For the frozen LLM that answers the pathology knowledge query, we use Claude-3.5-Sonnet. Both visual and textual features are encoded using CONCH \cite{lu2024visual}, with a feature dimension $d$ of $512$ for both modalities. Key hyperparameters of FOCUS include: a window size $w$ of $32$ (Sec. \ref{sec:3.2.1}), a compression ratio $\gamma$ of $0.8$ (Sec. \ref{sec:3.2.2}), and a similarity threshold $\theta_{\text{base}}$ of $0.7$ (Sec. \ref{sec:3.3}).
The model is optimized using AdamW with a learning rate of $1\times10^{-4}$ and trained for up to $80$ epochs with early stopping, selecting the best checkpoint based on validation performance.

\noindent\textbf{Evaluation Metrics.} We employ three evaluation metrics: balanced accuracy (Balanced ACC), F1 score, and the area under the curve (AUC). To ensure robust evaluation, we conduct ten-fold cross-validation, where each fold contains a unique sampling of train/validation/test data. The mean and standard deviation across all folds are reported. 

\subsection{FSWL Results}\label{sec:4.2} To evaluate the effectiveness of FOCUS, we compare our framework against SOTA methods in general pathology image analysis approaches including TransMIL \cite{shao2021transmil}, DS-MIL \cite{li2021dual}, DTFD-MIL \cite{zhang2022dtfd}, and WiKG-MIL \cite{li2024dynamic}, as well as SOTA methods in few-shot pathology image analysis like TOP-MIL \cite{qu2024rise} and ViLa-MIL \cite{shi2024vila}. Additionally, we establish strong baselines using foundation model features with Max Pooling, Mean Pooling, and Attentive Pooling as linear probing methods. CONCH \cite{lu2024visual} is applied to extract visual features for all methods. We compare FOCUS with these methods across three datasets under three few-shot settings ($4$-shot, $8$-shot, and $16$-shot), as shown in Table \ref{tab:results_16}. 

\noindent\textbf{TCGA-NSCLC (2 classes).} FOCUS achieves the best performance across all metrics and shot settings on this dataset. Notably, in the $4$-shot setting, our method achieves a Balanced ACC of $81.9\%$, outperforming the second-best method (ViLa-MIL) by $1.2\%$. The improvement is particularly significant in AUC, where FOCUS reaches $91.5\%$, surpassing ViLa-MIL's $88.5\%$ by a margin of $3\%$. This performance advantage is maintained in higher-shot settings, with FOCUS achieving $97.2\%$ AUC in the 16-shot scenario, demonstrating that our method can match or exceed the performance of fully-supervised approaches while requiring only a fraction of the labeled training data.

\noindent\textbf{CAMELYON (2 classes).} FOCUS shows remarkable improvement in low-shot scenarios on this dataset. In the challenging $4$-shot setting, our method achieves an ACC of $70.1\%$, significantly outperforming all baseline methods, with the next best method ViLa-MIL achieving only $65.8\%$. As the number of shots increases, FOCUS maintains competitive performance, achieving the best AUC scores in both $8$-shot ($85.6\%$) and $16$-shot ($94.3\%$) settings.

\noindent\textbf{UBC-OCEAN (5 classes).} FOCUS shows strong performance across all few-shot settings on this more complex dataset. In the $4$-shot scenario, our method achieves the highest ACC ($70.4\%$) and AUC ($91.1\%$), significantly outperforming other methods. This advantage persists in higher-shot settings, with FOCUS achieving state-of-the-art AUC of $96.7\%$ in the $16$-shot setting, surpassing the previous best (DS- and TOP-MIL tied at $95.6\%$) by $1.1\%$.

The extensive experimental results across three pathology datasets demonstrate that FOCUS consistently outperforms previous SOTA methods, especially in the most challenging $4$-shot scenarios, in which our method achieves absolute improvements of $1.2\%$, $4.3\%$, and $3.3\%$ in Balanced ACC over the second-best methods on the TCGA-NSCLC, CAMELYON, and UBC-OCEAN datasets, respectively. The performance advantages are maintained as the number of available shots increases, with FOCUS achieving the highest AUC scores of $97.2\%$, $94.3\%$, and $96.7\%$ in the $16$-shot setting across the three datasets.

\subsection{Ablation Studies}\label{sec:4.3}
\subsubsection{Effects of Key Modules in FOCUS}
To evaluate the contribution of each module within FOCUS to the overall performance, we conduct an ablation study on the UBC-OCEAN dataset using $5$ variants under $4$-, $8$-, and $16$-shot settings. The experimental results, measured by Balanced ACC with ten-fold cross-validation, are reported in Table \ref{tab:ablation_modules}.
\begingroup
\setlength{\tabcolsep}{4pt} 
\renewcommand{\arraystretch}{1.5} 
\begin{table}[ht]
\centering
\caption{Ablation study of key modules of FOCUS on UBC-OCEAN under $4$-, $8$-, and $16$-shot settings (via Balanced ACC).}
\begin{adjustbox}{width=8.2cm}
\scalebox{0.8}{
\begin{tabular}{cccccc|c}
\toprule
 & BaseMIL & Prompt & KAVTC & SVTC& CrossAgg~~& Balanced ACC \\
\midrule
\multirow{5}{*}{\rotatebox[origin=c]{90}{\textbf{4-shot}}} 
&   $\checkmark$ & & & &  & $0.623_{\pm 0.086}$  \\
&   $\checkmark$& $\checkmark$& & &  & $0.638_{\pm 0.074}$ \\
&  $\checkmark$ & $\checkmark$& $\checkmark$& &  & $0.673_{\pm 0.082}$ \\
&   $\checkmark$& $\checkmark$& $\checkmark$& $\checkmark$&  & $0.688_{\pm 0.056}$  \\
&   $\checkmark$& $\checkmark$& $\checkmark$& $\checkmark$& $\checkmark$ & $\bm{0.704_{\pm 0.088}}$ \\
\midrule
\multirow{5}{*}{\rotatebox[origin=c]{90}{\textbf{8-shot}}} 
&   $\checkmark$ & & & &  & $0.682_{\pm 0.034}$ \\
&   $\checkmark$& $\checkmark$& & &  & $0.698_{\pm 0.049}$  \\
&  $\checkmark$ & $\checkmark$& $\checkmark$& &  & $0.731_{\pm 0.037}$   \\
&   $\checkmark$& $\checkmark$& $\checkmark$& $\checkmark$&  & $0.769_{\pm 0.045}$   \\
&   $\checkmark$& $\checkmark$& $\checkmark$& $\checkmark$& $\checkmark$ & $\bm{0.773_{\pm 0.054}}$ \\
\midrule
\multirow{5}{*}{\rotatebox[origin=c]{90}{\textbf{16-shot}}} 
&   $\checkmark$ & & & &  & $0.779_{\pm 0.040}$  \\
&   $\checkmark$& $\checkmark$& & &  & $0.798_{\pm 0.038}$  \\
&  $\checkmark$ & $\checkmark$& $\checkmark$& &  & $0.827_{\pm 0.030}$   \\
&   $\checkmark$& $\checkmark$& $\checkmark$& $\checkmark$&  & $0.849_{\pm 0.048}$  \\
&   $\checkmark$& $\checkmark$& $\checkmark$& $\checkmark$& $\checkmark$ & $\bm{0.864_{\pm 0.032}}$\\
\bottomrule
\end{tabular}}
\end{adjustbox}
\label{tab:ablation_modules}
\end{table}
\endgroup

\noindent We systematically analyze these variants by first establishing a baseline \textbf{BaseMIL} that performs only patch feature aggregation. We then incrementally added components: the \textbf{Prompt} variant incorporates a language prompt branch and computes similarity scores between slide-level features and class-specific text features for prediction. Next, we progressively integrated the knowledge-enhanced adaptive visual token compression (\textbf{KAVTC}, Figure \ref{fig:methodology}(a)), sequential visual token compression (\textbf{SVTC}, Figure \ref{fig:methodology}(b)), and cross-modal aggregation (\textbf{CrossAgg}, Figure \ref{fig:methodology}(c)) modules. The experimental results demonstrate consistent performance improvements with each additional component, indicating the effectiveness of our proposed three-stage progressive and adaptive visual compression strategy.

\subsubsection{Effects of Pathology FM Encoders}
\begin{figure}[htbp]
  \centering
  \includegraphics[width=8.2cm]{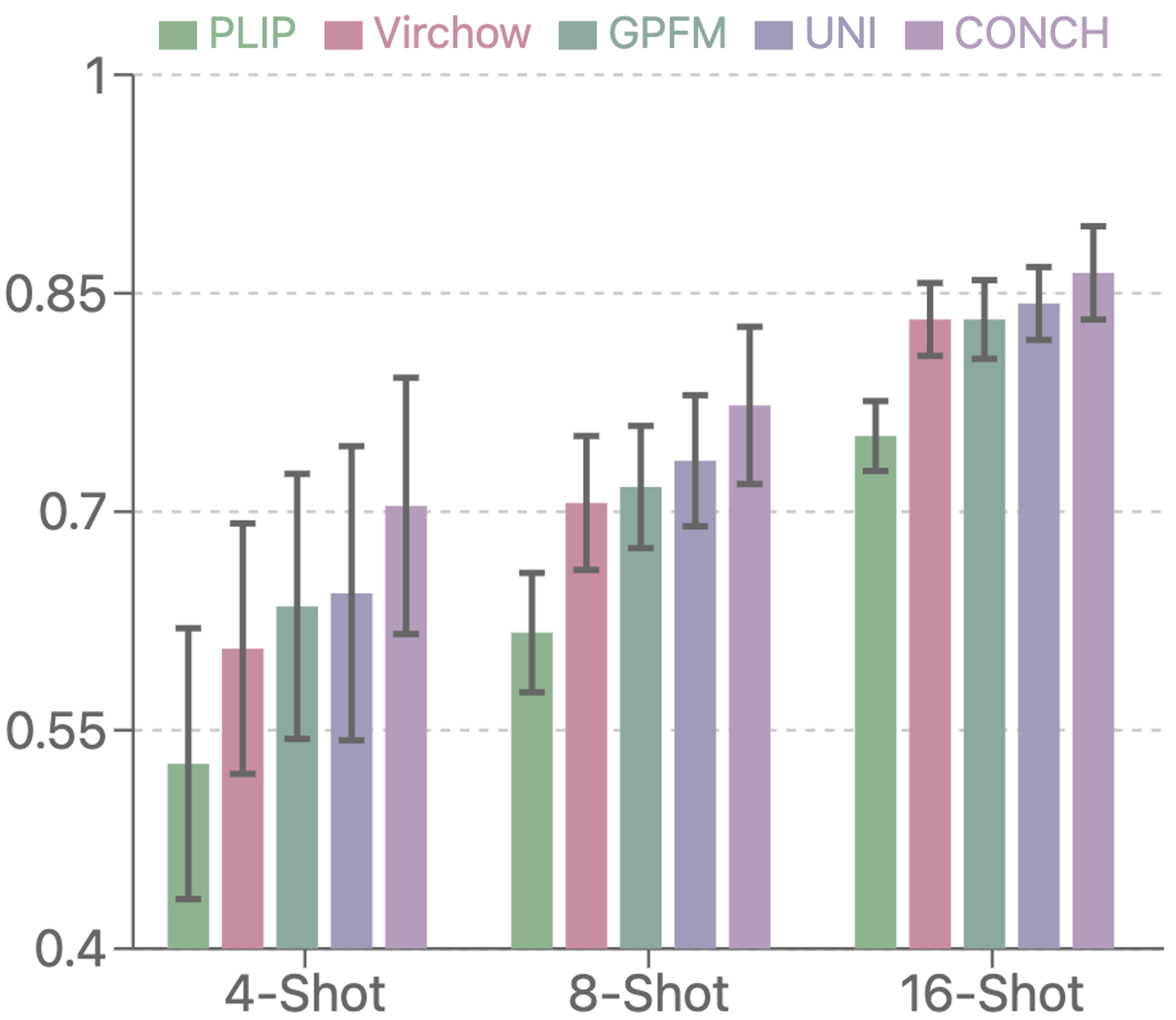}
  \caption{Performance comparison of different foundation models (UNI, GPFM, Virchow, PLIP, and CONCH) on UBC-OCEAN under few-shot settings (measured by Balanced ACC).} 
  \vspace{-0.5cm}
  \label{fig:ablation_fm}
\end{figure}

To evaluate the effect of using different pathology FM as feature extractors for FOCUS, we employ 5 pathology FMs, \textit{i.e.}, UNI \cite{chen2024towards}, GPFM \cite{ma2024towards},  Virchow \cite{vorontsov2024foundation}, PLIP \cite{huang2023visual}, and CONCH \cite{lu2024visual} for comparison. Figure \ref{fig:ablation_fm} shows the implications of different FMs as feature extractors across various few-shot settings, measured by Balanced ACC via ten-fold cross-validation. CONCH consistently outperforms other FMs, achieving the highest Balanced ACC across all settings ($70.4\%$, $77.3\%$, and $86.4\%$ for $4$-shot, $8$-shot, and $16$-shot, respectively). This can be attributed to the fact that CONCH is pre-trained with large-scale pathological image-caption pairs, potentially generating more aligned visual and textual features. Interestingly, despite being a vision-language model, PLIP yields the lowest performance among both multi-modal and uni-modal FMs. Besides, all FMs show substantial improvement as the number of shots increases, with the most significant gains observed in $4$- and $8$-shot settings. The performance gap between FMs narrows under $16$-shots, with UNI ($84.3\%$), GPFM ($83.2\%$), and Virchow ($83.2\%$) achieving comparable results. 


\subsubsection{Effects of LLMs for Prompt Generation}\label{sec:4.3.3}
\begin{figure}[htbp]
  \centering
  \includegraphics[width=8.5cm]{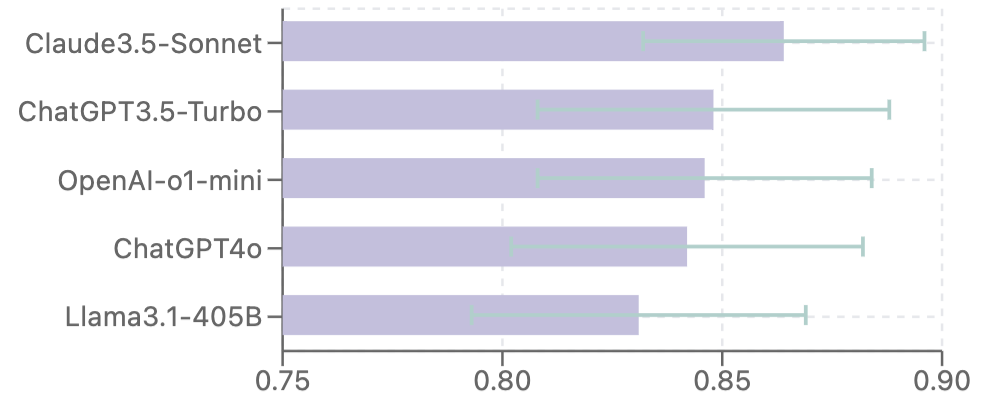}
  \caption{Performance comparison of prompts from different LLMs on UBC-OCEAN under 16-shot settings (Balanced ACC).} 
  \vspace{-0.2cm}
  \label{fig:ablation_llm}
\end{figure}

We also explore the effects of employing different LLMs for pathology knowledge query answering and the candidates include ChatGPT3.5-Turbo, ChatGPT-4o \cite{achiam2023gpt}, OpenAI-o1-mini, Claude-3.5-Sonnet, and Llama3.1-405B \cite{dubey2024llama}. Each LLM is prompted with identical instructions (detailed in Supplementary Material) to generate the visual descriptions used in FOCUS. We conduct this ablation study on the UBC-OCEAN dataset under $16$-shots, with the results shown in Figure \ref{fig:ablation_llm}. The experimental results demonstrate the strong capability of LLMs in generating effective pathology description prompts, with all variants achieving accuracies above $83\%$. Among the evaluated models, Claude3.5-Sonnet achieved the highest Balanced ACC of $86.4\%$, followed by ChatGPT3.5-Turbo ($84.8\%$) and OpenAI-o1-mini ($84.6\%$). These observations suggest that LLMs with enhanced language modeling capabilities and domain-specific knowledge can potentially further boost the performance of FOCUS.
\vspace{-0.2cm}

\section{Conclusion}\label{sec:conclusion}
This work presents FOCUS, a knowledge-enhanced adaptive visual compression multi-instance learning framework for few-shot WSI analysis. We propose a three-stage progressive and adaptive visual compression strategy that leverages pathology foundation models and domain-specific prompts to achieve visual redundancy removal and informative region prioritization. FOCUS operates without requiring burdensome inputs: it first conducts global redundancy removal through coarse-grained filtering based on pathology foundation models, integrates language descriptions for guided visual token compression by computing semantic relevance, then performs neighbor-aware visual token compression via fine-grained filtering, and finally a cross-modal aggregation module is applied to generate slide-level representation. Extensive experiments against SOTA methods on three pathology datasets across multiple cancer types demonstrate the effectiveness of FOCUS, showcasing strong few-shot weakly-supervised learning capability for WSI analysis and pushing the boundaries of CPath in data-scarce scenarios.
\section*{Acknowledgment}
This work was supported by the National Natural Science Foundation of China (No. 62202403), Hong Kong Innovation and Technology Commission (Project No. MHP/002/22 and ITCPD/17-9), Shenzhen Science and Technology Innovation Committee Fund (Project No. KCXFZ20230731094059008) and Research Grants Council of the Hong Kong Special Administrative Region, China (Project No. R6003-22 and C4024-22GF).

{
    \small
    \bibliographystyle{ieeenat_fullname}
    \bibliography{main}
}

\clearpage
\renewcommand\thesection{\Alph{section}}
\setcounter{section}{0}
\maketitlesupplementary

\section{Pathology Prior Knowledge Prompt}
This section lists the language prompt generated by LLMs for each dataset, including TCGA-NSCLC, CAMELYON, and UBC-OCEAN (see Section \ref{tcga}, \ref{camelyon}, and \ref{ubc-ocean}). For TCGA-NSCLC and CAMELYON, Claude-3.5-Sonnet is used for prior knowledge generation. While for UBC-OCEAN, Claude-3.5-Sonnet, ChatGPT3.5-Turbo, ChatGPT-4o, OpenAI-o1-mini, and Llama3.1-405B are used, as mentioned in the main paper. Moreover, in Section \ref{analysis}, we include a detailed analysis of different LLMs' prompts for potential reasons that lead to the performance ranking shown in the ablation studies of the main paper (Section \ref{sec:4.3.3}).

\subsection{TCGA-NSCLC}\label{tcga}
\textbf{Lung adenocarcinoma (LUAD):} A whole slide image of lung adenocarcinoma at high resolution with visually descriptive characteristics of clear cytoplasm, round or oval nuclei, prominent nucleoli, rich vascularity, irregular blood vessels, intratumoral septa, and heterogeneity.

\noindent\textbf{Lung squamous cell carcinoma (LUSC):} A whole slide image of lung squamous cell carcinoma at high resolution with visually descriptive characteristics of squamous cell differentiation, round structures with eosinophilic cytoplasm, distinct cell borders and abundant cytoplasm, enlarged nuclei, irregular nuclear shape, increased chromatin density.

\subsection{CAMELYON}\label{camelyon}
\textbf{Normal:} Normal lymph node, high resolution: Uniform small lymphocytes densely packed. Well-formed follicles in cortex, lymphocyte cords in medulla. Thin-walled blood vessels throughout. No atypical cells or architectural distortions.

\noindent\textbf{Breast cancer metastases:} Metastatic lymph node, high resolution: Large, pleomorphic cancer cells interspersed in lymphoid tissue. Cells show irregular nuclei, prominent nucleoli, abundant cytoplasm. Atypical cell arrangements. Desmoplasia, increased mitoses, potential necrosis. Abnormal blood vessels present.

\subsection{UBC-OCEAN}\label{ubc-ocean}
\subsubsection{Claude-3.5-Sonnet}
\textbf{Clear-cell ovarian carcinoma:} A whole slide image of clear-cell ovarian carcinoma at high resolution reveals large polygonal cells with clear cytoplasm and distinct cell boundaries. Nuclei are hyperchromatic and pleomorphic. Characteristic "hobnail" cells with nuclei bulging into cystic spaces are visible.

\noindent\textbf{High-grade serous carcinoma:} A whole slide image of high-grade serous carcinoma at high resolution displays marked nuclear atypia with high nuclear-to-cytoplasmic ratio. Nuclei are highly pleomorphic with prominent nucleoli. Numerous mitotic figures and areas of necrosis are present.

\noindent\textbf{Low-grade serous carcinoma:} A whole slide image of low-grade serous carcinoma at high resolution exhibits uniform population of cells with mild to moderate nuclear atypia. Psammoma bodies (calcified concretions) are often visible.

\noindent\textbf{Endometrioid carcinoma:} A whole slide image of endometrioid carcinoma at high resolution shows back-to-back glands lined by columnar cells with stratified nuclei. Squamous differentiation may be present. Nuclear atypia varies with tumor grade.

\noindent\textbf{Mucinous carcinoma:} A whole slide image of mucinous carcinoma at high resolution exhibits tall columnar cells with abundant intracellular mucin, often with goblet cell features. Nuclei are usually basally located. Degree of nuclear atypia varies with tumor grade.

\subsubsection{ChatGPT-3.5Turbo}
\textbf{Clear-cell ovarian carcinoma:} Clear-cell ovarian carcinoma typically presents with hobnail cells containing clear cytoplasm and distinct cell borders.

\noindent\textbf{High-grade serous carcinoma:} High-grade serous carcinoma displays complex papillary structures lined by atypical cells with high nuclear to cytoplasmic ratio and prominent nucleoli.

\noindent\textbf{Low-grade serous carcinoma:} Low-grade serous carcinoma demonstrates small glands or papillae lined by cells with low-grade nuclear features and scant cytoplasm.

\noindent\textbf{Endometrioid carcinoma:} Endometrioid carcinoma exhibits glandular differentiation with stratification, nuclear atypia, and architectural complexity resembling endometrial tissue.

\noindent\textbf{Mucinous carcinoma:} Mucinous carcinoma is characterized by pools of extracellular mucin and tall columnar cells with basally located nuclei and apical mucin.

\subsubsection{ChatGPT-4o}
\textbf{Clear-cell ovarian carcinoma:} Clear-cell ovarian carcinoma: Characterized by large, clear cells with abundant cytoplasm and distinct cell borders, often forming tubulocystic, papillary, or solid patterns.

\noindent\textbf{High-grade serous carcinoma:} High-grade serous carcinoma: Noted for its high-grade nuclear atypia, frequent mitotic figures, and complex papillary structures with slit-like spaces.

\noindent\textbf{Low-grade serous carcinoma:} Low-grade serous carcinoma: Features small, uniform cells with mild nuclear atypia, forming papillary and micropapillary structures with less mitotic activity.

\noindent\textbf{Endometrioid carcinoma:} Endometrioid carcinoma: Resembles endometrial tissue with glandular patterns, often with squamous differentiation and varying degrees of nuclear atypia.

\noindent\textbf{Mucinous carcinoma:} Mucinous carcinoma: Composed of tall columnar cells with abundant intracellular mucin, forming glandular or cystic structures with varying degrees of atypia.

\subsubsection{OpenAI-o1-mini}
\textbf{Clear-cell ovarian carcinoma:} Clear-cell ovarian carcinoma: Characterized by cells with clear cytoplasm and distinct cell borders, often exhibiting hobnail nuclei and sharply defined cell outlines under high-resolution imaging.

\noindent\textbf{High-grade serous carcinoma:} High-grade serous carcinoma: Features densely packed, irregularly shaped cells with prominent nucleoli, high nuclear-to-cytoplasmic ratios, frequent mitotic figures, and abundant psammoma bodies on high-resolution analysis.

\noindent\textbf{Low-grade serous carcinoma:} Low-grade serous carcinoma: Presents with uniform, mildly atypical cells arranged in papillae or glands, minimal mitotic activity, and a more cohesive architecture when examined at high resolution.

\noindent\textbf{Endometrioid carcinoma:} Endometrioid carcinoma: Displays glandular structures resembling endometrial tissue, with tubular formations, moderate cytologic atypia, and occasional squamous differentiation under high-magnification views.

\noindent\textbf{Mucinous carcinoma:} Mucinous carcinoma: Identified by large mucin-filled vacuoles within the cells, abundant extracellular mucin, and expansive, often multiloculated glandular structures observable in high-resolution microscopy.

\subsubsection{Llama3.1-405B}
\textbf{Clear-cell ovarian carcinoma:} Clear-cell ovarian carcinoma: Characterized by hobnail or clear cells with abundant clear cytoplasm, often with a tubulocystic or solid growth pattern and a network of delicate, branching vessels.

\noindent\textbf{High-grade serous carcinoma:} High-grade serous carcinoma: Exhibits a complex, branching architecture with slit-like spaces and a high nuclear-to-cytoplasmic ratio, featuring large, pleomorphic nuclei with prominent nucleoli and frequent mitoses.

\noindent\textbf{Low-grade serous carcinoma:} Low-grade serous carcinoma: Displays a predominantly micropapillary or cribriform growth pattern with small, uniform nuclei and a low mitotic rate, often with a characteristic microinvasive pattern of invasion.

\noindent\textbf{Endometrioid carcinoma:} Endometrioid carcinoma: Comprises glandular structures resembling endometrial glands, with columnar or cuboidal cells having oval nuclei and a variable amount of eosinophilic cytoplasm, often with squamous differentiation.

\noindent\textbf{Mucinous carcinoma:} Mucinous carcinoma: Characterized by large, mucin-filled glands or cysts lined by tall, columnar cells with basally located nuclei and a prominent apical mucin droplet, often with a characteristic intestinal-type epithelial differentiation.

\begin{algorithm*}[ht]
\caption{LAViC-MIL: Language-guided Adaptive Visual Compression Multi-instance Learning}
\label{alg:lavic-mil}
\small
\begin{algorithmic}[1]
\Require Training set of whole slide images $\{X^l\}_{l=1}^L$ with labels $\{Y_l\}_{l=1}^L$ ($Y_l \in \{1,2,\ldots,S\}$), pathology description prompts $\mathbf{T^P}$, window size $w$, base similarity threshold $\theta_{base}$, compression ratio $\gamma$, maximum sequence length $M_{max}$, maximum epochs $E_{max}$
\Ensure Optimal model parameters $\Theta^*$ of LAViC-MIL that minimize $\mathcal{L} = -\log P(Y_l|\mathbf{B}^l_c, \mathbf{T})$

\State Initialize model parameters $\Theta$
\For{epoch $e = 1$ to $E_{max}$}
    \For{each WSI $X^l$ in training set}
        \State Segment patches $X^l = \{x^l_1, ..., x^l_{N_l}\}$ from tissue regions
        \State Extract patch features via pathology foundation models $\mathbf{B}^l = \{f(x^l_i) : x^l_i \in X^l\} = \{\mathbf{b}^l_i\}^{N_l}_{i=1}$
        \State $\mathbf{T} = [\mathbf{T^L}; \mathbf{T^P}]$ \Comment{Concatenate learnable and pathology prompts}

        \State // Stage 1: Global Redundancy Removal via Pathology FMs
        \For{each window of size $w$ in $\mathbf{B}^l$}
            \State $\mathbf{\hat{b}}^l_i = \mathbf{b}^l_i/\|\mathbf{b}^l_i\|_2$ \Comment{Normalize features}
            \State $S_{ij} = \mathbf{\hat{b}}^l_i \cdot \mathbf{\hat{b}}^l_j$ \Comment{Pairwise similarities}
            \State $\tau_g = \mu(S) + \sigma(S)$ \Comment{Dynamic threshold}
            \State $R_i = \frac{1}{w}\sum^w_{j=1} S_{ij}$ \Comment{Mean similarity}
        \EndFor
        \State Remove patches where $R_i > \tau_g$

        \State // Stage 2: Language-guided Visual Token Prioritization
        \State $\mathbf{A} = \text{softmax}(\frac{(\mathbf{T}W_q)(\mathbf{B}^lW_k)^\top}{\sqrt{d}})$,\quad$r_i = \frac{1}{t_1 + t_2}\sum^{t_1+t_2}_{j=1} \mathbf{A}_{ji}$ \Comment{Token relevance}
        \State $k = \min(M_{max}, \gamma N'_l)$ \Comment{Number of tokens to select}
        \State $\mathbf{B}^l_s = \{b^l_i|\text{rank}(r_i) \leq k\}$ \Comment{Select top-k tokens}

        \State // Stage 3: Sequential Visual Token Compression
        \For{stage $i$ with threshold $\theta_i = \theta_{base} + i\Delta\theta$}
            \State $s_{j,j+1} = \frac{\mathbf{b}^{l\top}_j \mathbf{b}^l_{j+1}}{\|\mathbf{b}^l_j\|_2 \cdot \|\mathbf{b}^l_{j+1}\|_2}$ \Comment{Cosine similarity}
            \State $\text{mask}_j = \begin{cases} 1, & \text{if } \min(s_{j-1,j}, s_{j,j+1}) < \theta_i \\ 0, & \text{otherwise} \end{cases}$
            \State $\mathbf{B}^l_{i+1} = \{\mathbf{b}^l_j \in \mathbf{B}^l_i : \text{mask}_j = 1\}$
        \EndFor

        \State // After Compression: Cross-modal Aggregation and Loss Computation
        \State $\text{Head}_i = \text{softmax}(\frac{\mathbf{Q}W^i_q(\mathbf{K}W^i_k)^\top}{\sqrt{d_k}})\mathbf{V}W^i_v$
        \State $\mathbf{O} = \text{LayerNorm}(\text{Concat}(\text{Head}_1, ..., \text{Head}_h)W_o)$
        \State $P(Y_l|\mathbf{B}^l_c, \mathbf{T}) = \text{softmax}(W_c\mathbf{O} + \beta_c)$
        \State $\mathcal{L}_{\text{CE}} = -\log P(Y_l|\mathbf{B}^l_c, \mathbf{T})$
        \State Update parameters $\Theta$ using gradient descent on $\mathcal{L}_{\text{CE}}$ via AdamW optimizer
    \EndFor
\EndFor
\State \Return Optimal model parameters $\Theta^*$
\end{algorithmic}
\end{algorithm*}

\subsection{Prompt Analysis for Different LLMs}\label{analysis}
As shown in Fig. \ref{fig:ablation_llm}, the superior performance of Claude-3.5-Sonnet ($86.4\%$ Balanced ACC) can be attributed to its comprehensive and clinically precise descriptions. Its prompts consistently balance cellular, architectural, and nuclear features while maintaining a clear focus on primary diagnostic criteria. For example, in describing clear-cell ovarian carcinoma, it precisely details both cellular characteristics (``large polygonal cells with clear cytoplasm") and distinctive features (``hobnail cells with nuclei bulging into cystic spaces"), providing rich, diagnostically relevant information that helps the model identify key discriminative features even with limited training examples.

ChatGPT3.5-Turbo achieved the second-best performance ($84.8\%$) with its more concise yet accurate descriptions. While less detailed than Claude's prompts, it successfully captures the essential diagnostic features of each cancer subtype. Its descriptions maintain clinical accuracy while focusing on primary diagnostic features, though they show slightly less consistency in description depth across different subtypes. This balance between conciseness and accuracy appears to provide sufficient guidance for the few-shot learning task.

OpenAI-o1-mini's prompts ($84.6\%$) strike a reasonable balance between detail and clarity, explicitly contextualizing features within high-resolution imaging. However, its descriptions occasionally include less clinically relevant details and tend to be more verbose without adding substantive diagnostic information. This might explain why its performance, while strong, falls slightly below that of more focused prompts.

The lower performance of ChatGPT-4o can be attributed to its overly brief descriptions that sometimes miss critical diagnostic features. Its prompts, while accurate, lack the depth and specificity needed for optimal few-shot learning. The brevity of its descriptions might not provide sufficient guidance for the model to distinguish between similar cancer subtypes, particularly in cases where subtle differences are diagnostically important.

Llama3.1-405B's prompts showed the lowest performance, likely due to their tendency to mix critical and non-critical features while using overly complex language. Its descriptions often include rare or less reliable diagnostic features and lack consistent structure across different cancer subtypes. For example, its inclusion of features like ``characteristic microinvasive pattern of invasion" might distract from more reliable primary diagnostic criteria, potentially confusing the model in the few-shot learning context.

\noindent\textbf{Intuitions and guidelines:} This analysis suggests that optimal prompts for few-shot pathology image classification should be comprehensive yet focused, maintaining a consistent structure while emphasizing established diagnostic criteria. The balance achieved by Claude-3.5-Sonnet between detail and clinical relevance appears to provide the most effective guidance for the model to learn discriminative features from limited training examples.

\end{document}